\documentclass[10pt]{cai}

\usepackage[table]{xcolor}

\usepackage{microtype}
\usepackage{float} 
\usepackage{graphicx} 
\usepackage{subcaption} 
\usepackage{multirow} 
\usepackage{diagbox} 
\usepackage{color,soul} 
\usepackage{linguex} 
\usepackage{svg} 
\usepackage{tikz} 
\usepackage{xcolor} 
\usepackage[T1]{fontenc}
\usepackage{amssymb}   
\usepackage{hyperref}
\usepackage{linguex} 

\usepackage{booktabs} 
\usepackage{array}    
\usepackage{caption}  

\definecolor{new_blue}{rgb}{0,0,1}
\definecolor{new_orange}{rgb}{0.91, 0.41, 0.17}
\definecolor{new_green}{rgb}{0.35, 0.69, 0.19}

\usepackage{listings}

\lstdefinelanguage{json}{
    basicstyle=\ttfamily\footnotesize,
    numbers=none,  
    showstringspaces=false,
    breaklines=true,
    frame=single
    }

\begin{document}
\def\conferenceyear{2025}
\volumeheader{38}{}{}
\begin{center}

\title{On the Influence of Discourse Relations in Persuasive Texts}
\maketitle

\thispagestyle{empty}

\begin{tabular}{cc}
Nawar Turk\upstairs{\affilone,*}, 
Sevag Kaspar\upstairs{\affilone}, Leila Kosseim\upstairs{\affilone}
\\[0.25ex]
{\small \upstairs{\affilone} Dept. of Computer Science and Software Engineering, Concordia University} \\
\end{tabular}
  
\emails{
  \upstairs{*}nawar.turk@mail.concordia.ca
}
\end{center}

\begin{center}
\textit{Published in: Proceedings of the 38th Canadian Conference on Artificial Intelligence (CanAI 2025), Calgary, Alberta, May 26–27, 2025.\\
Available at: \url{https://caiac.pubpub.org/canadian-ai-2025-long-papers}}

\end{center}

\vspace{0.2in}

\begin{abstract}
     This paper investigates the relationship between Persuasion Techniques (PTs) and Discourse Relations (DRs) by leveraging Large Language Models (LLMs) and prompt engineering. Since no dataset annotated with both PTs and DRs exists, we took the SemEval 2023 Task~3 dataset labelled with~19 PTs as a starting point and developed LLM-based classifiers to label each instance of the dataset with one of the~22~PDTB~3.0 level-2 DRs. In total, four LLMs were evaluated using~10 different prompts, resulting in~40 unique DR classifiers. Ensemble models using different majority-pooling strategies were used to create~5 silver datasets of instances labelled with both persuasion techniques and level-2 PDTB senses. The silver dataset sizes vary from 1,281 instances to 204 instances, depending on the majority pooling technique used. Statistical analysis of these silver datasets shows that six discourse relations (namely \textsc{Cause, Purpose, Contrast, Cause+Belief, Concession,} and \textsc{Condition}) play a crucial role in persuasive texts, especially in the use of \textit{Loaded Language, Exaggeration/Minimisation}, \textit{Repetition} and to cast \textit{Doubt}.
     This insight can contribute to detecting online propaganda and misinformation, as well as our general understanding of effective communication.

 All code and data developed for this work are publicly available at \url{https://github.com/CLaC-Lab/Persuasion-Discourse-Mapping}.
 
\end{abstract}

\copyrightnotice

\section{Introduction}

Persuasion techniques play a crucial role in communication and are often used to spread propaganda.  Due to their significance in shaping public opinion and today's ease at generating text via LLMs, a growing body of work in Natural Language Processing (NLP) has studied persuasive texts computationally and has developed inventories of specific writing techniques such as \textit{Appeal to Authority} and \textit{Loaded Language} to create more persuasive narratives.

On the other hand, the field of computational discourse analysis has a long history of studying the coherence structure of discourse by characterizing how text spans are connected logically to make a text coherent. In this view, discourse relations can be seen as general in nature as they characterize texts with varying communicative goals, whereas persuasion techniques characterize the specific relations used for the purpose of persuasion.

This paper investigates the relationship between PTs and DRs. To do so, we first developed LLM-based discourse relation classifiers capable of labelling text spans with PDTB 3.0 level-2 senses. We then applied ensembles of the top classifiers to label the human-labelled PT SemEval 2023 Task~3 English corpus (\cite{piskorski-etal-2023-semeval}) to create a series of~5 silver datasets with both DR and PT annotations. These silver corpora were then mined to uncover patterns between discourse relations and persuasive language. 

The main contributions of this paper include: 
\begin{enumerate}
    \item A comparative evaluation of various LLM-based models for annotating level-2 PDTB senses. 
    \item The design and release of silver corpora with both persuasion techniques and discourse relations annotations. 
    \item The identification of the discourse relations that are more often used to implement specific persuasion techniques in texts. 
    \end{enumerate}
This insight can help improve both the automatic identification of persuasive techniques and our broader understanding of effective communication to help combat online propaganda and misinformation.

\section{Background}

Persuasion techniques are essential for communication, but are often used to disseminate online propaganda. Driven by the increasing ease of text generation through LLMs, a growing body of work in NLP research has addressed the study of persuasion techniques, especially in the context of automatic detection of propaganda (e.g.~\cite{kiesel-etal-2019-semeval,da-san-martino-etal-2020-semeval,piskorski-etal-2023-semeval,dimitrov-etal-2024-semeval}).  Early work focused solely on binary document-level classification (propaganda or not), then~\cite{rashkin-etal-2017-truth} extended the classification to 4 specific types  (\textit{trusted}, \textit{satire}, \textit{hoax}, and \textit{propaganda}).  Later models addressed the detection of more fine-grained persuasion techniques within shorter textual spans (e.g.~\cite{kiesel-etal-2019-semeval,da-san-martino-etal-2020-semeval,piskorski-etal-2023-semeval,dimitrov-etal-2024-semeval}). Recent work has also extended the detection of persuasion techniques to multimodal media such as memes~\cite{dimitrov-etal-2024-semeval,nayak2024memept}. Parallel to this, several inventories of persuasion techniques have been developed based on rhetorical and psychological techniques. Such inventories include the 6-label fallacy inventory of~\cite{goffredo-etal-2023-argument}, the European Commission inventory~\cite{EU-per}, and the related inventory of~\cite{dimitrov-etal-2021-semeval}.  This latter inventory is currently the most widely used  as it was adopted by the SemEval 2021, 2023 and 2024 Shared Tasks on the detection of persuasion techniques~\cite{piskorski-etal-2023-semeval,dimitrov-etal-2024-semeval}. This list defines around 20 persuasion techniques such as \textit{causal oversimplification}, \textit{name calling} and  \textit{smear}.  Example~\ref{ex:oversimplification} shows  a sentence labelled with a \textit{causal oversimplification} from the SemEval 2023 Task~3 definition\footnote{\href{https://propaganda.math.unipd.it/semeval2024task4/definitions.html}{https://propaganda.math.unipd.it/semeval2024task4/definitions.html}}.

\ex.
\begin{tabular}{lp{4.5in}}
 \texttt{"text":} & \texttt{The reason New Orleans was hit so hard with the hurricane was because of all the immoral people who live there. } \\
    \texttt{"PT":} & \texttt{causal oversimplification}\\
    \end{tabular}

\label{ex:oversimplification} 

Similar to persuasion techniques, discourse relations characterize the connection
between text spans. However, discourse relations focus on the more general structure of a text by indicating how text spans are connected logically to make a text coherent.
Discourse analysis has a longer history of research in the field of NLP, and several linguistic frameworks have been proposed to model discourse relations\footnote{Discourse relations are also known as ``coherence relations'' or ``discourse senses''.}. The most widely used frameworks are the Rhetorical Structure Theory (RST)~\cite{mann1988rhetorical} and the Penn Discourse Treebank (PDTB)~\cite{miltsakaki2004penn,prasad2007penn}. Following these frameworks, several annotated corpora have been developed, such as the RST-DT~\cite{corpus-rst-dt} and~3 versions of the Penn Discourse TreeBank (PDTB)~\cite{prasad2006penn,prasad2007penn,PDTB3AnnotationManual}. 
The PDTB takes into account discourse connectives (e.g. \textit{but, because,~\ldots}) and labels these cues with a discourse relation, called discourse {\em sense}. The most recent version of the PDTB (PDTB~3.0~\cite{PDTB3AnnotationManual})  organizes these senses in a 3-tier hierarchy:  level-1 which includes~4 basic senses (e.g. \textsc{Temporal, Contingency}), level-2 which includes~22 more specific senses (e.g. \textsc{Cause, Condition, Purpose}), and level-3 which includes very fine-grained senses.  Most work in the literature has used either level-1 or level-2 senses. 
Using the PDTB level-2 senses, Example~\ref{ex:oversimplification} would be annotated with a \textsc{Cause} discourse relation as in Example~\ref{ex:Cause}.

\ex.
\begin{tabular}{lp{4.5in}}
 \texttt{"text":} & \texttt{The reason New Orleans was hit so hard with the hurricane was because of all the immoral people who live there.} \\
    \texttt{"DR":} & \texttt{Cause}\\
    \end{tabular}
    \label{ex:Cause}

 Recent work has investigated the interplay between discourse structures and persuasive language. \cite{lei-huang-2023-discourse, chernyavskiy-etal-2024-unleashing}  have examined the relationship between discourse relations and  propaganda. \cite{lei-huang-2023-discourse} explored the connection between the four top-level PDTB discourse relations ({\sc Comparison}, {\sc Contingency}, {\sc Temporal}, and {\sc Expansion}) and the eight subtypes of news discourse structure defined in~\cite{choubey-etal-2020-discourse} with the presence of propaganda. Similarly, \cite{chernyavskiy-etal-2024-unleashing} examined the correlation between persuasion techniques and RST rhetorical relations. Additionally,~\cite{rehbein-2019-role} investigated the distribution of discourse connectives and 3 PDTB level-2 relations across various persuasive texts (talks, interviews, and articles); however, no specific persuasion technique was considered. To complement previous work, the goal of our paper is to systematically investigate the mapping between all level-2 PDTB discourse relations, and all persuasion techniques found in the SemEval 2023 dataset. 

\section{Overall Methodology}

\begin{figure}
\centering
{\includegraphics[scale=0.37]{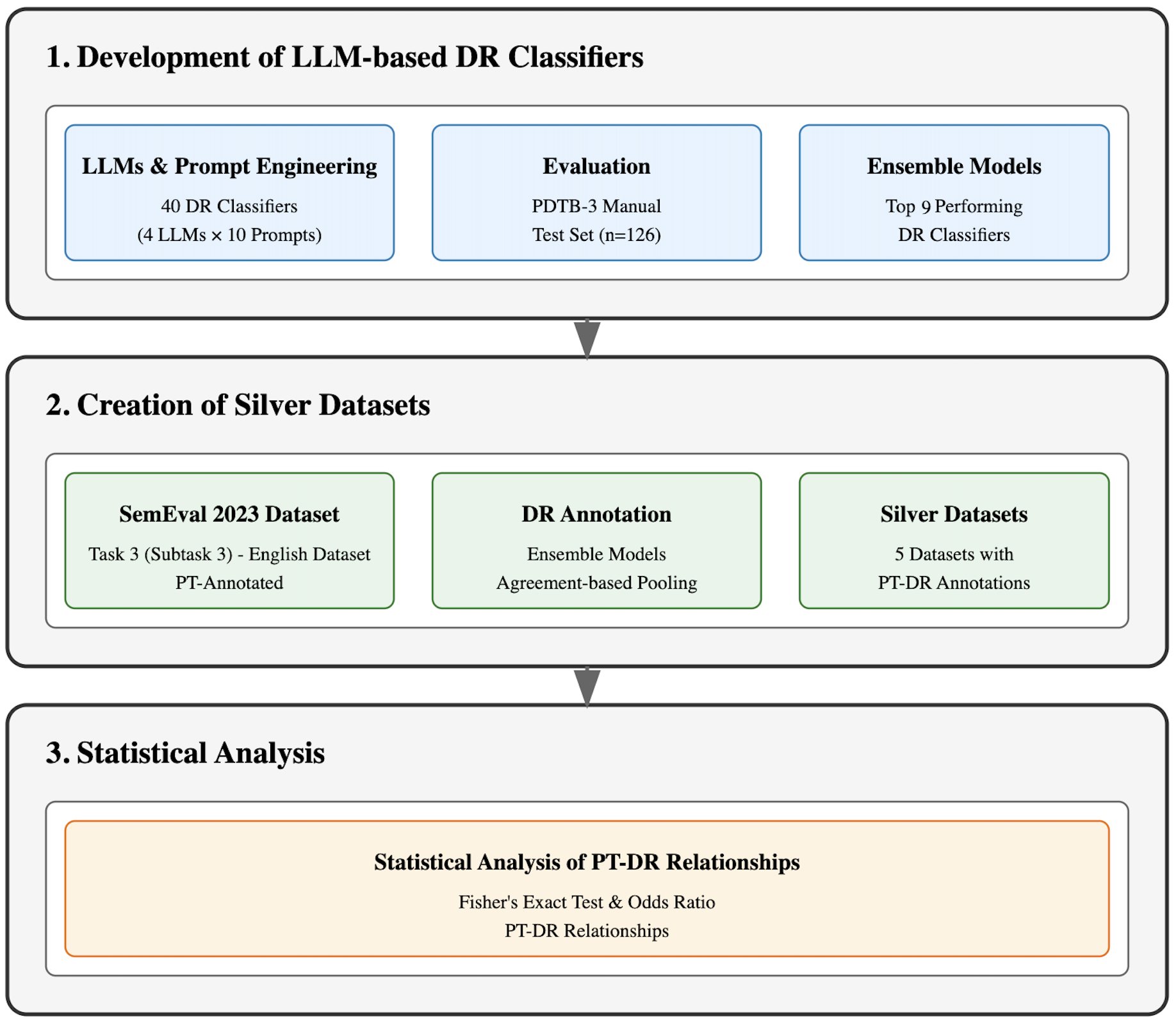}}
\caption{Overall Methodology}
\label{fig:system_Architecture_for_PT-DR_Association_Analysis} 
\end{figure}

To explore the relationship between persuasion techniques (PTs) and discourse relations (DRs), we needed a dataset annotated with both types of information. Since no such dataset was found, we created a series of 5 silver datasets. We chose the SemEval 2023 Task~3 English dataset~\cite{piskorski-etal-2023-semeval} that has been manually labelled with PTs as a starting point and annotated it automatically with DRs. We chose to automatically label DRs as opposed to starting with a DR-annotated dataset and labelling it with PTs for a few reasons. First, since significantly more research has been done on discourse parsing and PDTB than on persuasion techniques, we expected  LLMs to be more knowledgeable about identifying PDTB senses than about identifying PTs.  In addition, because the PDTB senses are organized hierarchically, this allowed us to label instances at different levels of granularity should the DR labelling be ambiguous. 

To annotate the SemEval 2023 Task~3 dataset, we considered using open-source PDTB parsers (e.g. \cite{Lin2014} and  \cite{Wang2015}); however, we experimented with LLMs and prompt engineering given their success in a variety of NLP-related tasks, and more recently in discourse parsing (\cite{thompson-etal-2024-llamipa,maekawa-etal-2024-obtain}).

 As shown in Figure~\ref{fig:system_Architecture_for_PT-DR_Association_Analysis}, the overall methodology consists of 3 main steps:
\begin{description}
\item [Development of LLM-based DR Classifiers] First, we developed discourse relation classifiers using LLMs and prompt engineering. We experimented with 4 LLMs and 10 prompts for a total of 40 configurations.  After evaluating them on a manually-annotated PDTB test set, we kept the top 9 best-performing classifiers to create an ensemble model. This is described in Section~\ref{sec:classifiers}.

\item [Creation of Silver Datasets] The top-performing DR classifiers were then used to annotate the SemEval 2023 Task~3~\cite{piskorski-etal-2023-semeval} with discourse relations. Using various agreement-based pooling strategies, we created 5 silver datasets annotated with both DRs and PTs with varying sizes and annotation confidence levels. This is described in Section~\ref{sec:silver}.

\item [Statistical Analysis between PTs and DRs] Finally, the silver datasets were used to conduct statistical analyses to identify the relationship between persuasion techniques and discourse relations. Our analysis revealed both strong associations between specific PTs and DRs, particularly with causal and contrastive relations. 
 This is described in Section~\ref{sec:analysis}. 
\end{description}

\section{LLM-based Discourse Relations Classifiers}
\label{sec:classifiers}
The task of the LLM-based DR classifiers is to predict one of the 22 level-2 senses from the PDTB 3.0  for each input instance. 

\subsection{Models and Prompt Design}

To develop the DR classifiers, we experimented with 4 different LLMs and 10 prompts, resulting in 40 DR classifiers\footnote{We call a DR classifier the combination of a specific LLM with a specific prompt.}. The four LLMs are \texttt{gpt-4o}, \texttt{Gemini 1.5-pro}, \texttt{Gemini 2.0-flash-exp}, and \texttt{Claude} (version \texttt{3.5-Sonnet-20241022}). Each LLM was queried using 10 prompts developed based on a variety of design elements~\cite{GeminiAPIDoc2023}, such as zero-shot instructions, few-shot examples, definitions, and chain-of-thought reasoning.  Table~\ref{tab:prompt-details} shows the description of these prompts. Although all prompts were designed to predict a PDTB level-2 sense, prompts 1 to 5 asked directly to identify a level-2 sense, while prompts 6 to 10 followed a chain-of-thought reasoning, requiring the LLM to first segment the given instance into two discourse arguments, identify the level-3 sense before determining the level-2 parent sense. The prompts also varied with respect to the supporting information they provided, such as whether they provided definitions (at level-2, level-3, or both) and examples (at level-2 or level-3)\footnote{All prompts and their specifications are publicly available at \url{https://github.com/CLaC-Lab/Persuasion-Discourse-Mapping}.}. Figure~\ref{fig:prompt3-excerpt} shows an example of Prompt~3, which provides the PDTB level-2 senses, their definitions and an example.

\begin{figure}[ht]
\centering
\scalebox{1}{
\begin{tabular}{|lp{4.6in}|} \hline
\texttt{"instruction":} & \texttt{Analyze the following paragraph and identify the most appropriate PDTB Level~2 discourse relation label from the list below. Use the definitions and examples provided to guide your decision. Only return the label with no additional explanation.} \\ 
&\\
\texttt{"paragraph":} & \texttt{\{paragraph\}} \\ 
&\\
\texttt{} & \texttt{**Available Labels with Definitions and Examples:**} \\ \texttt{"label":} & \texttt{Synchronous} \\ 
\texttt{"definition":} & \texttt{Temporal overlap between events.} \\ 
\texttt{"example":} & \texttt{"While the cake was baking, she prepared the icing."} \\ 
&\\
\texttt{"label":} & \texttt{Asynchronous} \\ 
\texttt{"definition":} & \texttt{One event precedes another.} \\ 
\texttt{"example":} & \texttt{"First, he packed his bags. Then, he left for the airport."} \\ 
&\\
\texttt{"label":} & \texttt{Cause} \\ 
\texttt{"definition":} & \texttt{Causal influence without a conditional relation.} \\ 
\texttt{"example":} & \texttt{"The roads were icy, thus causing the accident."} \\ 
&\\
\texttt{"label":} & \texttt{Cause+Belief} \\ 
\texttt{"definition":} & \texttt{Evidence provided to induce belief.} \\ 
\texttt{"example":} & \texttt{"The footprints in the snow led her to conclude someone had been there."} \\ 
&\\
\texttt{[...]} & \texttt{} \\\hline
\end{tabular}
}
\caption{Excerpt from Prompt 3 showing four PDTB level 2 senses with definitions and examples. [...] refers to the description of the remaining PDTB level 2 senses.}
\label{fig:prompt3-excerpt}
\end{figure}

\begin{table}
\centering
\begin{tabular}{|>{\centering\arraybackslash}m{2cm}|c|c|c|c|}
\hline
\textbf{Prompt ID} & \textbf{Sense level } & \textbf{Definition } & \textbf{Examples} & \textbf{Chain-of-thought} \\
 & \textbf{asked for} & \textbf{provided for} & \textbf{provided for} & \textbf{reasoning} \\
\hline
1 & level-2 & n/a & n/a & no \\
\hline
2 & level-2 & level-2 & n/a & no \\
\hline
3 & level-2 & level-2 & level-2 & no \\
\hline
4 & level-2 & level-2 & level-2 & no \\
\hline
5 & level-2 & level-2, level-3 & n/a & no \\
\hline
6 & level-3, level-2 & level-3 & n/a & yes \\
\hline
7 & level-3, level-2 & level-3 & n/a & yes \\
\hline
8 & level-3, level-2 & level-2, level-3 & n/a & yes \\
\hline
9 & level-3, level-2 & level-2, level-3 & n/a & yes \\
\hline
10 & level-3, level-2 & level-3 & level-3 & yes \\
\hline
\end{tabular}
\caption{Prompt Descriptions for {the LLM-Based Discourse Relation Classifiers}}
\label{tab:prompt-details}
\end{table}

\subsubsection{Classifier Evaluation}

 To evaluate the 40 DR classifiers, we used a test set of 126 instances extracted manually from the PDTB 3.0 Annotation Manual~\cite{PDTB3AnnotationManual} with their gold standard senses. The nuances of instances were chosen to include at least one PDTB level-3 sense, ensuring that all level-2 senses were covered. To be representative of an actual text, the dataset includes both explicit and implicit relations. We used this dataset as opposed to a larger one to reduce the cost of the classifier evaluation.

To evaluate the classifiers, each LLM+prompt combination was run twice, and the results were averaged.
We assessed their performance on the test set using three metrics:
\begin{enumerate}
    \item  Macro F1 scores.
\item Hallucination rate: the percentage of non-valid level-2 sense predictions.  This includes non-PDTB senses, ``no relation'' responses, and level-1 senses. Level-3 sense predictions were not considered hallucinations as they were mapped to their level-2 parent sense.  Note that macro-F1 scores do not include hallucinations in their calculation, as these can easily be filtered out.
\item Prediction consistency: the percentage of identical predictions across the two independent runs. 
\end{enumerate}

\begin{table}
\centering
\label{tab:model-average-performance}
\resizebox{1\textwidth}{!}{%
\begin{tabular}{|>{\centering\arraybackslash}m{5cm}|c|c|c|}
\hline
\textbf{LLM} & \textbf{Macro F1} & \textbf{Hallucinations} & \textbf{Consistency} \\
\hline
\texttt{claude-3-5-sonnet-20241022} & 0.35 & 1.6~\% & 79.9~\% \\
\hline
\texttt{gemini-1.5-pro}             & 0.31 & 0.7~\% & 90.9~\% \\
\hline
\texttt{gemini-2.0-flash-exp}       & 0.30 & 0.1~\%  & 85.6~\% \\
\hline
\texttt{gpt-4o}                   & 0.23 & 0.2~\%  & 71.2~\% \\
\hline
\end{tabular}
}
\caption{Performance of the LLMs Across all 10 Prompts on the PDTB Test Set. Results are averaged across all prompts and runs per LLM.}
\label{tab:model-average-performance}
\end{table}

\begin{figure}
    \centering
    \includegraphics[width=\textwidth]{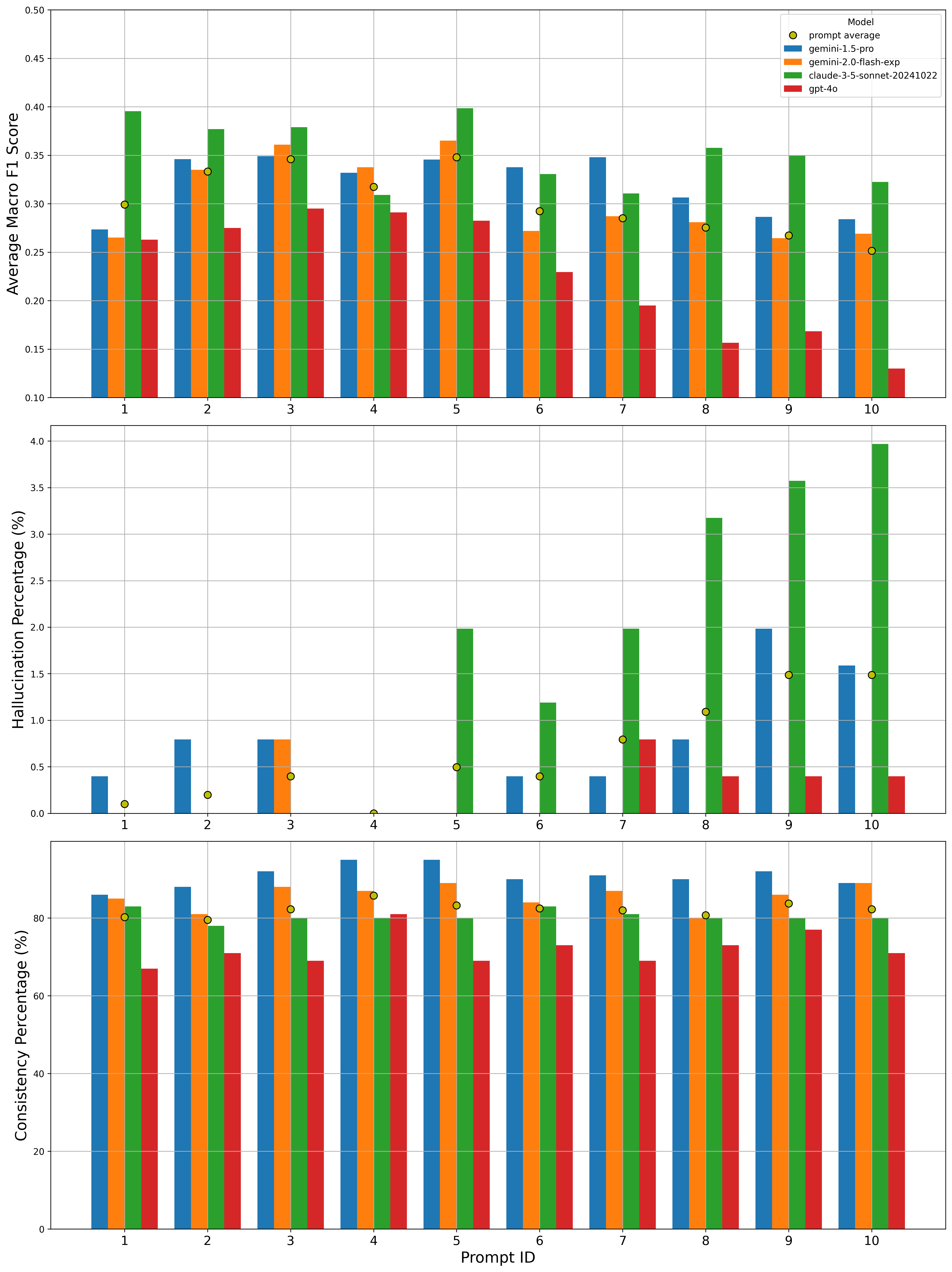}
    \caption{Impact of  Prompts on LLM Performance for DR Classification. Each bar represents the average of two independent runs per LLM on that prompt. Prompt average shows the mean performance across all four LLMs for each prompt.}
    \label{fig:parser-macro-f1}
\end{figure}

Table~\ref{tab:model-average-performance} shows a summary of the performance of each LLM; while Figure~\ref{fig:parser-macro-f1} shows details of each classifier. In Figure~\ref{fig:parser-macro-f1}, LLMs are grouped by the prompt they use (1 to 10), and each bar represents the average of the two independent runs. The average for each prompt across all LLMs is shown as a dot (between the orange and green bars). As Table~\ref{tab:model-average-performance} and Figure~\ref{fig:parser-macro-f1} show, \texttt{claude-3-5-sonnet-20241022} achieves the highest average macro F1 score of 0.35. However, this advantage is accompanied by the highest number of hallucinations {(1.6\%)} and moderate prediction consistency (79.9\%). In contrast, \texttt{gemini-1.5-pro} demonstrates strong overall performance; although its F1 score is slightly lower at 0.31, it exhibits the highest consistency rate of 90.9~\% and lower hallucinations, making it a reliable choice for stable performance. \texttt{gemini-2.0-flash-exp} also reports a macro F1 score of 0.30, similar to \texttt{gemini-1.5-pro}, but distinguishes itself by having the fewest hallucinations (0.1\%). Conversely, \texttt{gpt-4o}, despite making fewer hallucinations, demonstrates the lowest overall performance. 
Overall, these findings reveal the trade-offs between accuracy and reliability, but no individual classifier yields a strong enough performance to be used to annotate a dataset automatically.

\subsubsection{Ensemble Classifiers} \label{sec:ensemble} 
\begin{figure}
\centering
\captionsetup{skip=0pt} 
\includegraphics[width=3.5in]{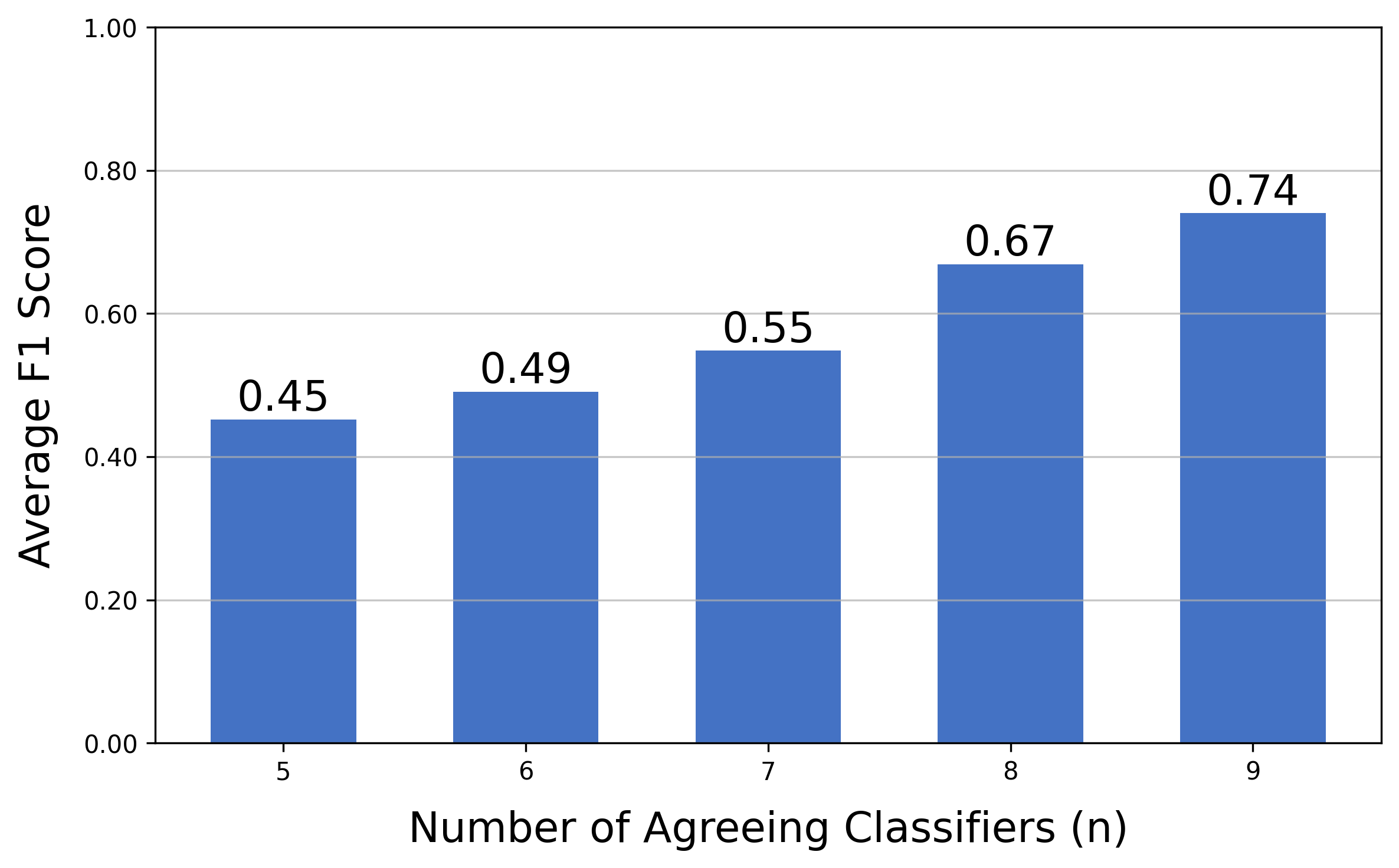}
\caption{Macro F1 Score of the Ensemble DR Classifiers}
\label{fig:pooling_f1} 
\end{figure}

Given that individual classifier performances were below expectations, we adopted an ensemble approach using majority voting. From the 40 classifiers, we selected the nine best-performing ones based on the highest macro F1 scores, strong prediction consistency between runs, and lower hallucination rates. These top-performing classifiers were used as an ensemble by pooling their predictions and taking the majority prediction. By imposing that at least $n$  classifiers ($n \in [5...9]$) agree on the prediction, we created 5 ensemble classifiers.
As shown in Figure~\ref{fig:pooling_f1}, the macro F1 scores of these 5 ensemble classifiers increase steadily from 0.45 (for $n=5$) to~0.74 (for $n=9$).  In addition, given the agreement constraint, none of the ensemble classifiers generated any hallucinations. Given these higher performances, we used these ensemble models to annotate discourse relations in the SemEval~2023 Task~3 (Subtask~3) dataset~\cite{piskorski-etal-2023-semeval}, thereby creating 5 silver datasets.

\section{Creation of Silver Datasets}
\label{sec:silver}

To create silver datasets with both persuasion techniques and discourse relations labels, we used the human-annotated SemEval 2023 Task 3 (Subtask 3)~\cite{piskorski-etal-2023-semeval} dataset annotated with PTs and used the ensemble-models to annotate it automatically with DRs. 
In order to use this dataset for DR labelling, we had to pre-process it. 

Since some instances were labelled with multiple PTs, we split multi-label entries into individual instances, with each instance labelled with a single PT. This  pre-processing  increased the dataset size from 3,760 to 5,853 instances. 

The dataset was extremely imbalanced, particularly for \textit{Loaded Language} and \textit{Name Calling/labelling}, which accounted for $30.9\%$ and $16.7\%$ of the instances, respectively.  To reduce the imbalance, we randomly undersampled the over-represented PTs. The reduced dataset is still imbalanced, but the top 2 PTs account for a more manageable $30.1\%$ of the instances, down from $47.6\%$. 


The next pre-processing step filtered out instances that contained more than 2 sentences. This was done because PDTB parsing typically focuses on identifying relations across arguments within a single sentence or across adjacent sentences.  To avoid cases where multiple DRs would be valid labels, we filtered out instances with more than 2 sentences. After this pre-processing step, the dataset was reduced to 2,064 instances.

\subsection{Silver Datasets}
We used the ensemble DR classifiers to annotate the processed SemEval PT dataset with PDTB~3.0 level-2 senses as DRs. As described in Section~\ref{sec:ensemble}, we experimented with 5 majority-agreement pooling strategies based on the agreement of $n$ classifiers out of~9 (where $n \in [5...9]$). The resulting silver datasets ranged in size from 1,281 instances for Silver-5 (i.e. $n=5$), 937 for Silver-6, 641 for Silver-7, 388 for Silver-8, and 204 for Silver-9. 
These five silver datasets show the inherent trade-off between dataset size and annotation confidence. 

\begin{figure}
\centering
\includegraphics[width=6in]{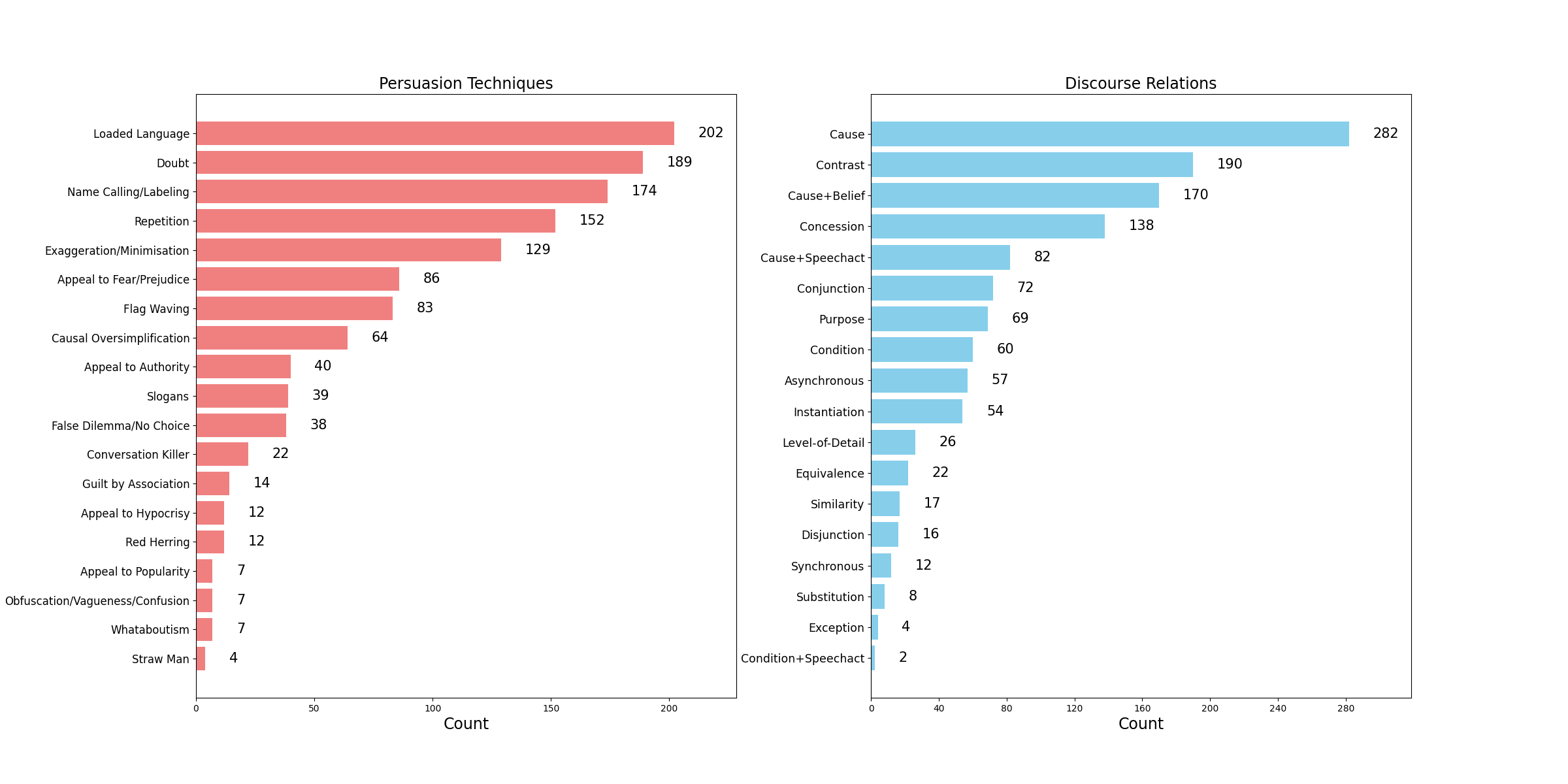}
\caption{Distribution of PTs and DRs in the Silver-5 Dataset ($1,281$ instances)}
\label{fig:Persuasion_Techniques_vs_Discourse_Relations_filtered} 
\end{figure}

Figure \ref{lst:dual-annotation-example} shows a dually annotated instance from the Silver-5 dataset, while Figure~\ref{fig:Persuasion_Techniques_vs_Discourse_Relations_filtered} shows the distribution of PTs and DRs in the same dataset.  As the figure shows, the distribution of DRs in  Silver-5 is imbalanced, with some DRs being significantly over-represented, but this imbalance is to be expected and also occurs in the entire PDTB~3.0 and many other discourse labeled corpora~\cite{REHBEIN16.457}.

\begin{figure}
\scalebox{1}{
\begin{tabular}{|lp{3.8in}|} \hline
 \texttt{"text":} &\texttt{"While we may never know the exact circumstances surrounding what transpired in the shooter’s hotel room, the information being released not just by law enforcement, but witnesses to the event who recorded hundreds of cumulative hours of video and audio, now calls the entirety official story into question.",}\\
\texttt{"PT"}: &\texttt{"Doubt",}\\
   \texttt{"DR"}: &\texttt{"Concession"}\\ \hline
\end{tabular}
}
    \caption{Example of a Dually Annotated Instance in a Silver Dataset.}
    \label{lst:dual-annotation-example}
\end{figure}

\section{Statistical Analysis of the Silver Datasets}
\label{sec:analysis}

The goal of the silver datasets was to explore the relationship between persuasion techniques and discourse relations. To have statistically significant results, our analysis focused on PTs and DRs that appear at least~25~times each in the silver datasets.

\begin{figure}[h]
\centering
\captionsetup{skip=0pt} 
\includegraphics[width=.9\linewidth]{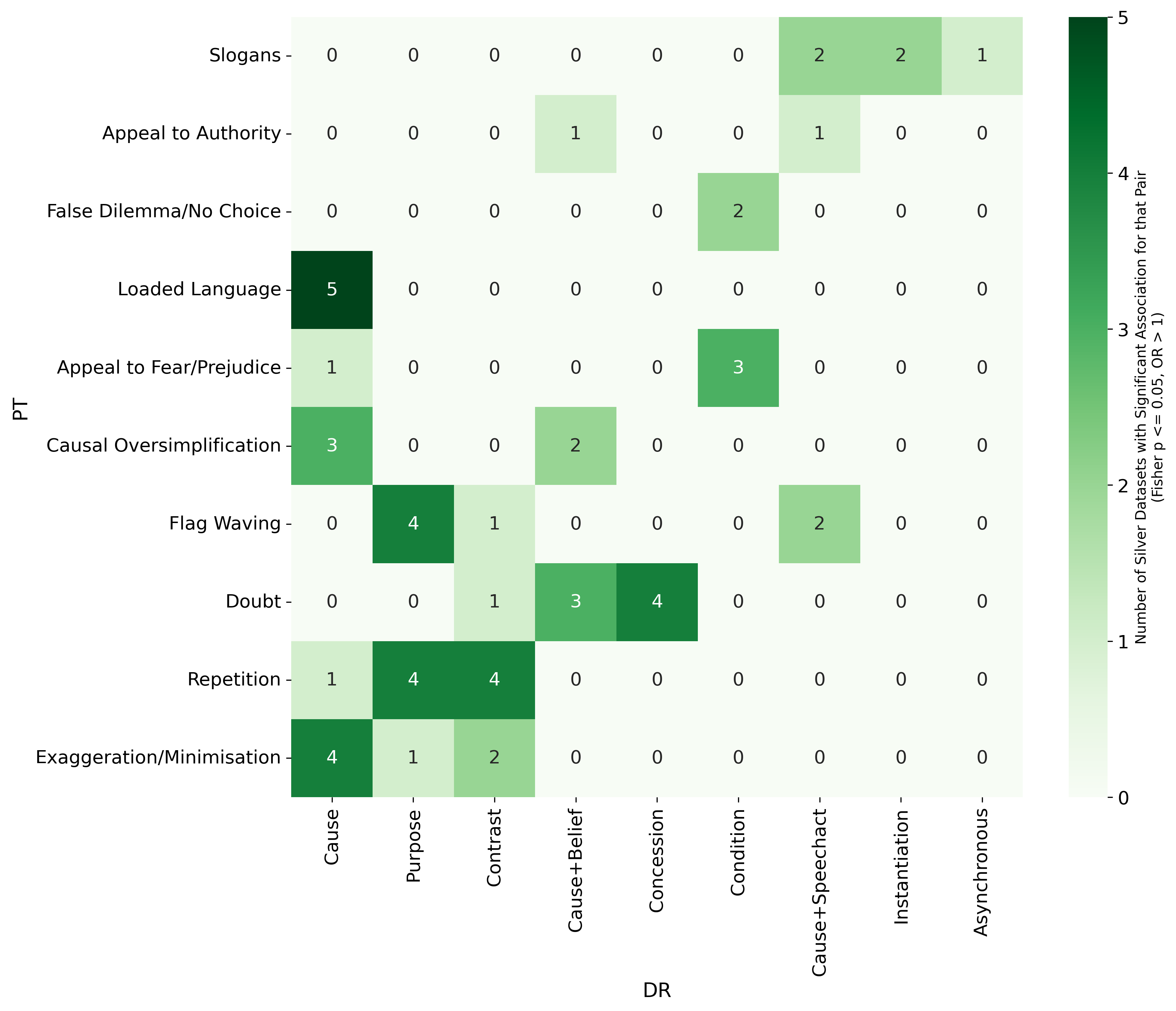}
\caption{Statistically Significant PT-DR Associations in the Silver Datasets (Fisher's exact test:  $p \leq 0.05$, $OR > 1$).}

\label{fig:Significant_PT-DR} 
\end{figure}

We used Fisher's exact test to identify statistically significant associations between DRs and PTs. Since some contingency tables of PT-DR pairs contain values less than~5, we used Fisher's exact test as it is known to be more accurate than the $\chi^2$ test for small sample sizes.  The odds ratio (OR) was then used to determine the direction of these associations, with values greater than~1~indicating positive associations. 
Figure~\ref{fig:Significant_PT-DR} shows a heat map of all PT-DR pairs and the number of silver datasets where the pair has a statistically significant positive association (i.e.~Fisher's $p \leq 0.05$ and OR~>~1). For example, the PT \textit{Loaded Language} has a statistically significant association with a \textsc{Cause} DR in all 5 silver datasets, while no dataset identified an association with a \textsc{Contrast} DR.
As Figure~\ref{fig:Significant_PT-DR} shows, several PT-DR associations clearly stand out. The PDTB senses of \textsc{Cause}, \textsc{Purpose}, and \textsc{Contrast} are the most frequently associated DRs, each paired with five, three, and four PTs, respectively.

Persuasion Techniques such as \textit{Loaded Language} and  \textit{Exaggeration/Minimisation} strongly correlate with the \textsc{Cause} discourse relation
while the \textit{Repetition} PT is consistently expressed through \textsc{Purpose} and \textsc{Contrast} discourse relations.
This suggests that persuasive texts often rely on causal reasoning, purposive statements, and contrasting ideas to have more impact. Similarly, the \textsc{Cause+Belief} discourse relation shows associations with multiple PTs, including \textit{Doubt} and \textit{Casual Oversimplification}. 

Notably, only 6 out of 22 senses showed significant correlations with persuasion techniques. The practical implication is that when writing persuasive text, focusing on these six discourse relations (\textsc{Cause, Purpose, Contrast, Cause+Belief, Concession,} and \textsc{Condition}) would likely lead to more effective persuasion.

\section{Conclusion and Future Work}

This paper explored the relationship between discourse relations (DRs) and persuasion techniques (PTs) through the use of Large Language Models (LLMs). 
 Several individual LLM-based DR classifiers were developed using a variety of prompt engineering methods.  Although no individual classifier reached a high enough performance to annotate a dataset with PDTB level-2 senses automatically, ensemble methods reached macro F1 ranging from 0.45 to 0.74. These ensemble models were then used to build silver corpora based on the SemEval 2023 Task~3 (Subtask~3)~\cite{piskorski-etal-2023-semeval} dataset with human-labelled PTs and automatic DRs. A statistical analysis of these silver datasets allowed us to identify pairs of PTs and DRs that co-occur more often, as well as six central discourse relations
 (\textsc{Cause, Purpose, Contrast, Cause+Belief, Concession,} and \textsc{Condition}) that play a crucial role in persuasive communication. 
This insight can advance the automatic detection of persuasive techniques and contribute to the detection of online propaganda and misinformation, as well as our general understanding of effective communication.

Future work includes improving the performance of the LLM-based classifiers in order to create larger and more precise silver datasets to allow for a deeper analysis. 
This work has leveraged a manually-annotated PT corpus (the SemEval 2023 Task 3 dataset) and labelled it with DRs automatically; however, the other direction should also be investigated -- i.e. leveraging a manually DR-annotated dataset, such as the PDTB~3.0~\cite{PDTB3AnnotationManual} with $\approx 46K$ relations, and building classifiers to label these instances with PTs. Also, this work has assumed that an instance from the dataset can be labelled with a single DR. We should investigate to what degree this assumption holds, as several DRs may be appropriate. Recent work by~\cite{costa2024multitask} supports this view by treating  implicit discourse relations identification as a multi-label task. Finally, this work has only investigated English texts, yet the SemEval 2023 Task 3 dataset does include annotations for a variety of languages (French, German, Italian, Polish, and Russian).  It would be interesting to explore to what extent the relation between PTs and DRs is language-independent or if different languages (or cultures) use discourse relations differently in persuasive communication.   

\section*{Acknowledgments}
The authors would like to thank the anonymous reviewers for their comments.
This work was financially supported by the Natural Sciences and Engineering Research Council of Canada (NSERC) and the Pierre Arbour Foundation.

\printbibliography[heading=subbibintoc]

\end{document}